\documentclass[journal]{IEEEtran}

\usepackage{graphicx}
\usepackage{amsmath, amsfonts, color, amssymb, mathrsfs, bm}
\usepackage[linesnumbered,ruled,vlined]{algorithm2e}
\usepackage{booktabs}
\usepackage{stfloats}
\usepackage{multirow}
\usepackage{float}
\usepackage{color,xcolor}
\usepackage{marvosym}
\usepackage{threeparttable}
\usepackage{cite}
\usepackage{hyperref}

\begin{document}
%
\title{RingFed: Reducing Communication Costs in Federated Learning on Non-IID Data}
%
%
%

\author{Guang~Yang\textsuperscript{1},
        Ke~Mu\textsuperscript{1},
        Chunhe~Song\textsuperscript{2},
        Zhijia~Yang\textsuperscript{2},
        and~Tierui~Gong\textsuperscript{3}
\thanks{\textsuperscript{1} University of Chinese Academy of Sciences, Beijing, China.  Correspondence to: Guang Yang \href{mailto:yangguang@sia.cn}{$\langle$yangguang@sia.cn$\rangle$}, Ke Mu\href{mailto:muke@sia.cn}{$\langle$muke@sia.cn$\rangle$}}

\thanks{\textsuperscript{2}Shenyang Institute of Automation, Chinese Academy of Sciences, Shenyang, China.  Correspondence to: Chunhe Song \href{mailto:songchunhe@sia.cn}{$\langle$songchunhe@sia.cn$\rangle$}, Zhijia Yang \href{mailto:yang@sia.ac.cn}{$\langle$yang@sia.ac.cn$\rangle$}}

\thanks{\textsuperscript{3}Institute of Computing Technology, Chinese Academy of Sciences, Beijing, China. Correspondence to: Tierui Gong \href{mailto:gongtierui@sia.cn}{$\langle$gongtierui@sia.cn$\rangle$}}
}

\maketitle

\begin{abstract}
	Federated learning is a widely used distributed deep learning framework that protects the privacy of each client by exchanging model parameters rather than raw data. However, federated learning suffers from high communication costs, as a considerable number of model parameters need to be transmitted many times during the training process, making the approach inefficient, especially when the communication network bandwidth is limited. This article proposes RingFed, a novel framework to reduce communication overhead during the training process of federated learning. Rather than transmitting parameters between the center server and each client, as in original federated learning, in the proposed RingFed, the updated parameters are transmitted between each client in turn, and only the final result is transmitted to the central server, thereby reducing the communication overhead substantially. After several local updates, clients first send their parameters to another proximal client, not to the center server directly, to preaggregate. Experiments on two different public datasets show that RingFed has fast convergence, high model accuracy, and low communication cost.
\end{abstract}


%
\IEEEpeerreviewmaketitle

\section{Introduction}
%
%
%
%
\IEEEPARstart{M}{achine} learning (ML), especially deep learning (DL), has made great breakthroughs in intelligent applications (e.g., image classification, image generation, object detection, and text translation \cite{Liu2018Event,Lecun2015Deep, Du2020Novel}), due to tremendous data generated by smart clients and the unprecedented computational capability of computers. Generally, an ML paradigm consists of two parts: training and inference, both of which are performed on a central server \cite{Li2017Multi} with powerful calculating ability and fast disk writing and reading, and all the original data are used to obtain an outstanding model. During inference, however, clients and servers that are physically far away will take a long time to transfer data, which is intolerable for scenarios with low-latency requirements \cite{Mao2017Survey}.  Therefore, the single central server is separated into several edge servers to process data close to the clients to reduce time delays \cite{Park2019Wireless}. Moreover, several advanced distributed learning optimization algorithms accelerate rapid training with multiple clients \cite{Lin2017Deep,Wang2018Edge}. Nonetheless, the performance of the edge server is not comparable to that of the central server and ultimately does not produce model results.

With the increase in the calculating ability of portable devices and people’s concern about personal data privacy, in 2017, Google proposed federated learning (FL), which is within the framework of distributed learning \cite{Mcmahan2017Communication}. 
In most cases, FL allows multiple clients to collaborate to solve a deep or machine learning problem under the coordination of a central server or service provider. Each client’s raw data are stored locally and not exchanged or transferred; instead, focused updates intended for immediate aggregation are used to achieve the learning objective. The lifecycle of FL is described below. First, FL can make full computational use of every client connected to the central server. The more clients that join, the more global the model behaves. Second, clients only upload parameters of models that are encrypted for the server to update the global model, protecting sensitive information of users and companies. Next, the central server aggregates parameters transmitted by clients with a specific algorithm to optimize the global model. Finally, clients download weights to update the model locally.

Although considerable effort has been put into FL research, several key challenges remain to be solved:
\begin{itemize}
	\item Communication costs. In FL, each client needs to communicate with the central server for global model updating. It is intolerable to create network congestion and make the total communication costs more expensive as the number of clients increases, especially when the communication network bandwidth is limited \cite{Mcmahan2017Communication}.
	\item Statistical heterogeneity. Unlike traditional datasets generated by specific organizations that are independent and identically distributed (IID), the dataset generated by each client in FL is non-IID overall because every client trains the model locally using only its own data. Therefore, it is difficult to make the global model converge \cite{Smith2017Federated,Li2018Federated}.
	\item System heterogeneity. The different hardware of the devices throughout the network can result in some devices running slowly, typically called stragglers, which are unable to complete the required number of training sessions in a period of time for the same model. These stragglers cause severe delay in the whole FL system
	 \cite{Li2018Near,Bonawitz2019Towards}.
	\item Encryption for privacy. In general, complex cryptographic algorithms make encryption more valid, but the training of the model already consumes a large quantity of computational resources. In addition, the complexity of the cryptographic algorithm depends on the additional trade-off on the client directly. Therefore, a good balance between complexity and effectiveness is needed \cite{Melis2019Exploiting}.
\end{itemize}


This article focuses on reducing communication costs. Additionally, the loss (accuracy) of the global model determined by the aggregation algorithm is equally important. The contributions of this paper are summarized as follows:
\begin{itemize}
	\item We introduce a new communication network structure. Compared to the traditional star topology, which decreases communication efficiency when updating parameters, a ring topology setup allows clients to pass information among each other, thereby improving the communication efficiency within a certain range.
	\item We propose a new optimization method based on the proposed RingFed. The new method takes advantage of the fact that clients can communicate with each other after a certain number of local training sessions. Instead of using the central server directly for aggregation, the new method performs preaggregation between clients, reducing communication costs and accelerating convergence..
	\item We demonstrate that the method can reduce communication costs significantly. Empirically, RingFed can achieve a lower loss value and faster convergence in fewer communication rounds with the CNN classifier on the MNIST and FMNIST datasets transformed to be non-IID.
\end{itemize}
The remainder of this paper is organized as follows. We introduce an overview of related work on FL to improve communication efficiency in Section \ref{section2}. Then, we present a basic model of FL and optimization methods, FedAvg, and our proposed RingFed in detail in Section \ref{section3}. In Section \ref{section4}, empirical simulation results for RingFed and three other outstanding algorithms are shown under various conditions, illustrating significant improvements in communication costs achieved by RingFed. Finally, we conclude our work in Section \ref{section5}.

\section{Related Work}\label{section2}

Communication is one of the main bottlenecks limiting the performance of FL. In the FL setting, a single communication round from each client may contain millions of parameters for complicated deep learning models \cite{He2016Deep}, resulting in high communication costs. Common strategies to improve communication efficiency can be categorized as updating schemes, model compression and end computation.

\textbf{Updating Schemes.} Updating Schemes. Some recent approaches have significantly improved upon traditional distributed learning methods \cite{Dekel2012Optimal}. These approaches propose more permissive means of local updating without requiring all devices to be involved in every communication round. In the convex setting, Smith \textit{et al.} \cite{Smith2017Federated} proposes MOCHA, which enables personalized FL for each client learning separated but related models by solving a primal-dual optimization borrowed from multitask learning. Smith \textit{et al.} \cite{Smith2018Cocoa} presents a general distributed learning framework, CoCoA, and extends it to cover general nonstrongly convex regularizers. In the nonconvex setting, FedAvg  \cite{Mcmahan2017Communication} averages parameters from each client after conducting local stochastic gradient descent (SGD), which works well according to experimental results. As the primal method of federated learning, FedAvg is difficult to tune and often exhibits unexpected divergence. FedNova \cite{Wang2020Tackling} eliminates objective inconsistency and preserves fast convergence by correctly normalizing local model updates when averaging. FedOpt \cite{Reddi2020Adaptive} extends adaptive optimizers, including AdaGrad, Adam, and Yogi of traditional deep learning, to FL, substantially improving the performance. SCAFFOLD \cite{Karimireddy2020Scaffold} uses the control variates to correct for “client drift” in its local updates. Moreover, other important local updating methods, such as local SGD, have been widely studied \cite{Lin2020Don, Stich2019Local, Wang2019Cooperative, Wang2019Adaptive}.

Instead of local updating, importance-based updating is also covered in \cite{Lim2020Federated}. Tao \textit{et al.} \cite{Tao2018Esgd} keeps track of the loss function values for each communication round. Uploading of clientside parameters that are important is performed only if the loss function value of the current round is less than that of the previous round. However, discarding these small values completely makes the global model converge slower and results in a loss of accuracy. Lin \textit{et al.} \cite{Lin2017Deep} aggregates the discarded values and performs parameter upload once when the set threshold value is reached.

\textbf{Model Compression.}
Reducing the number of transmission parameters and compressing the model is one way to decrease the communication consumption. Konevcny \textit{et al.} \cite{Konevcny2016Federated} proposes two ways to reduce upload costs: structured and sketched updates. proposes two ways to reduce upload costs: structured and sketched updates. By multiplying the two matrices together, structured updates force the parameter matrix of each client to change to a low-rank matrix. Sketched updates encode the client-side parameters before sending to reduce the parameter size and then decode them at the server to restore the original parameter data. Furthermore, Caldas \textit{et al.} \cite{Caldas2018Expanding} introduces lossy compression and federated dropout, which allow users to efficiently train locally on smaller subsets of the global model and provide a reduction in both client-to-server communication and local computation. Similarly, in each communication round, Khaled \textit{et al.} \cite{Khaled2019Gradient} compresses iterates using the lossy randomized compression technique, followed by the computation of the gradients.

\textbf{End Computation.}
Mcmahan \textit{et al.} \cite{Mcmahan2017Communication} proposes the FedAvg algorithm and finds that increasing the number of training epochs per client and the parallelisms of clients contributes to the performance of the global model. Similarly, Liu \textit{et al.} \cite{Liu2019Communication} proposes FedBCD, in which each client conducts diverse local updates first before uploading parameters. In \cite{Huang2020Loadaboost}, clients compare the cross-entropy loss from the current training round with the median loss from the previous round to decide whether to upload parameters, which is proven to accelerate the convergence rate of the global model. A two-stream model with maximum mean difference (MMD) constraints is used instead of a single model trained on the device, achieving a reduction of more than 20\% in required communication rounds \cite{Yao2018Two}. FedProx \cite{Li2018Federated} proposes adding a proximal term to each local objective function to make the algorithm more robust to the heterogeneity of the local objectives, improving the performance of federated averaging. Khaled \textit{et al.} \cite{Khaled2019First} assumes that all clients are involved, and batch gradient descent can result in faster convergence than stochastic gradients on clients.

\begin{figure*}[tp]
	\centering
	\includegraphics[scale=0.25, trim=0 0 0 0]{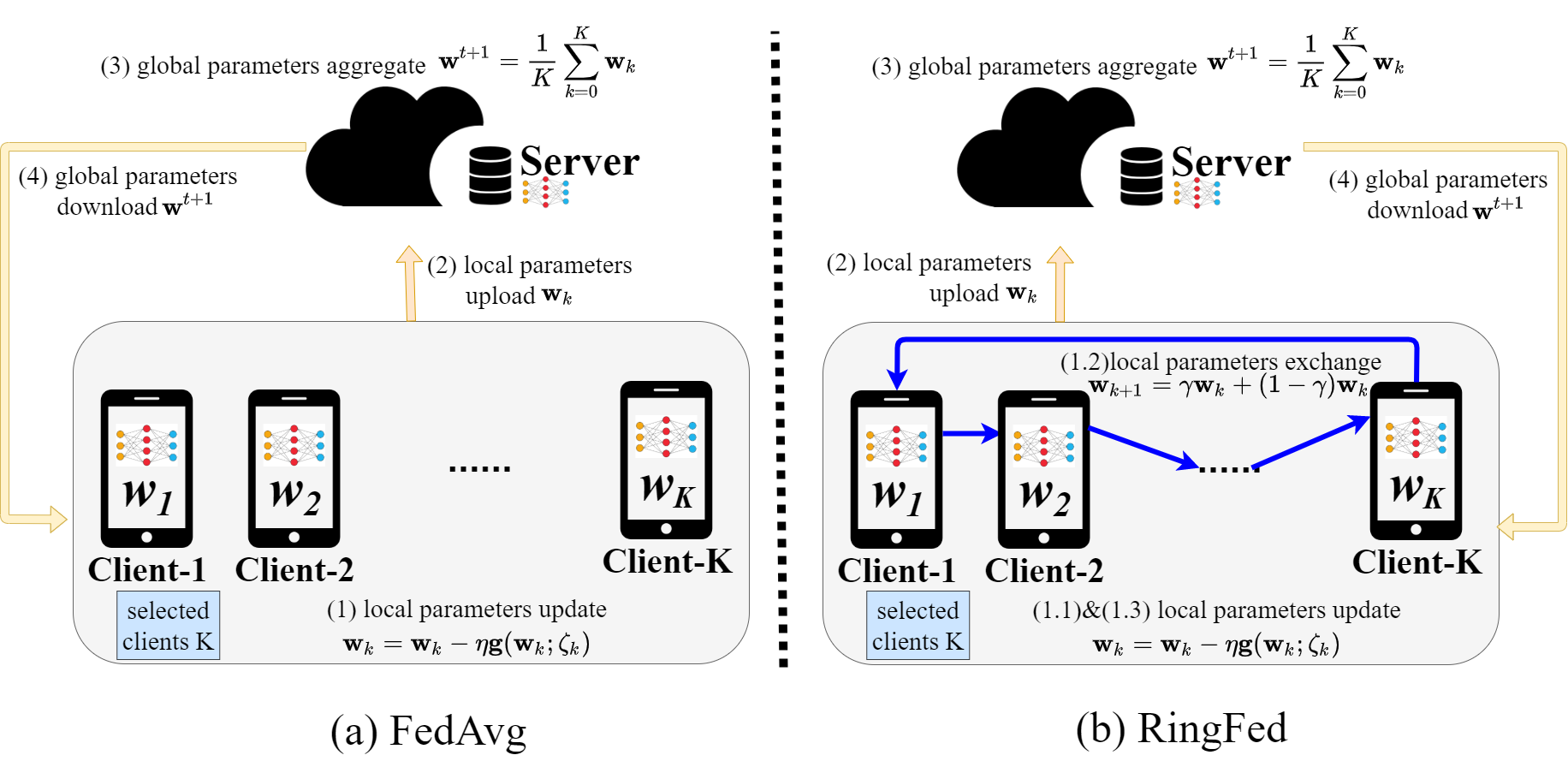}	
	\caption{Flow-charts of FedAvg and RingFed.}
	\label{fig:label_1}
\end{figure*}

\section{Optimization Methods}\label{section3}
In this section, we introduce more details about FL, focusing on the optimization methods of FedAvg and RingFed. FL is generalized distributed learning that makes all kinds of smart equipment contribute to developing an excellent intelligent model, which keeps data at each device and reduces delays of inference. We consider $\mathcal{K}$ clients in total, and the server randomly selects a certain percentage of $K$ ($K \subset \mathcal{K}$) clients for the training task in each communication round, i.e., the clients involved in the training task differ in each communication round. Formally, a local function is used for optimization at each client, which can be expressed as
\begin{equation}
	\begin{aligned}
		f_k(\mathbf{w}) = \sum_{k \in \mathcal{D}_k} \frac{1}{\left| \mathcal{D}_k \right|} F_k(\mathbf{w}),
		\label{LocalLoss}
	\end{aligned}
\end{equation}
where $f_k(\mathbf{w})=\mathcal{L}(\mathbf{x}_j,\mathbf{y}_j;\mathbf{w})$ denotes the loss function whose values indicate how well the model performs after one or more training epochs, $\mathbf{x}_j$ is the $j$-th input real vector consisting of the processed features, $\mathbf{y}_j$ is the corresponding label, and $\mathbf{w}$ is the parameters that construct the ML model. $\mathcal{D}_k$ is the training dataset generated by the $k$-th client locally and $\left| \cdot \right|$ denotes the cardinality of the dataset. Clearly, $\cup_{j=1}^{\left| \mathcal{D}_k \right|}(\mathbf{x}_j,\mathbf{y}_j)=\mathcal{D}_k$.
During an epoch $e$, each client $k$ runs SGD with:
\begin{equation}
	\begin{aligned}
		\mathbf{w}_k=\mathbf{w}_k-\eta \mathbf{g}(\mathbf{w}_k;\zeta_k),
		\label{SGD}
	\end{aligned}
\end{equation}
where $\eta$ is the learning rate, $\mathbf{g}(\mathbf{w}_k;\zeta_k)$ is the statistical gradient descent value of $F_k(\mathbf{w})$, and $\zeta_k$ denotes that the dataset $\mathcal{D}_k$ is sampled randomly at epoch $e$. In our work, we only focus on the DL loss function which means $F_k(\mathbf{w})$ is non-convex and the non-IID distribution of the dataset. We also assume that $\mathbb{E}_{\zeta_k\sim\mathcal{D}_k}[\mathbf{g}(\mathbf{w}_k;\zeta_k)]=\nabla F_k(\mathbf{w}_k)$, which we realize in experiments by setting codes properly.

The ultimate aim for FL is to minimize the global model, which can be expressed in finite-sum form as
\begin{equation}
	\begin{aligned}
		f(\mathbf{w}) = \sum_{k=0}^K \frac{\left| \mathcal{D}_k \right|}{\left| \mathcal{D} \right|} f_k(\mathbf{w}),
		\label{2}
	\end{aligned}
\end{equation}
where $\mathcal{D}$ denotes all the data of $K$ clients with $\cup_{i=1}^K\mathcal{D}_i=\mathcal{D}$, and all the data cannot be accessed by the server out of preserving privacy. 

\subsection{Federated Averaging (FedAvg)}
Due to the bottlenecks of limited network bandwidth and non-IID data in FL, FedAvg is proposed to reduce the number of communication rounds and mitigate the influence of non-IID datasets.

As illustrated in Fig. \ref{fig:label_1}, during an intact communication round of FedAvg, the central server first sends the parameters of the original global model to all clients. 
\begin{equation}
	\begin{aligned}
		\mathbf{w}=\mathbf{w}^t
		\label{DownloadParams}
	\end{aligned}
\end{equation}
Then, $k$ clients are selected by the central server randomly. Next, these clients train models locally with their own data for E epochs by SGD in \eqref{SGD}, after which the clients send parameters back to the central server. Finally, the central server averages all the uploaded parameters as an update of the global model in accordance with

\begin{equation}
	\begin{aligned}
		\mathbf{w}^{t+1}=\frac{1}{K}\sum_{k=0}^K\mathbf{w}_{k}
		\label{FedAvg}
	\end{aligned}
\end{equation}
The procedure of FedAvg is described in Algorithm \ref{alg1}.

In paper \cite{Mcmahan2017Communication}, the average of two models using two small subsets of the total dataset is proven to achieve significantly lower loss on the training dataset than the best model achieved by training on either of the small subsets.

However, the limitations of FedAvg restrict its performance, which requires further improvement. On the one hand, as the number of clients increases, the non-IID distribution of datasets overall is strengthened. It is difficult for FedAvg to fine-tune the hyperparameters (e.g., learning rate and number of layers) of neural network models if the overall dataset has a strong non-IID distribution, which leads to many more epochs required to make models converge. Furthermore, it is worse for some clients with low computational performance called stragglers not to complete extra training epochs to be abandoned in a communication round. On the other hand, a deeper model introduces an explosive number of parameters. Therefore, frequent communication with the central server to upload the substantial volume of packaged parameters can lead to long-term congestion, resulting in inefficient aggregation and slowing other important functions that require networking due to the communication network bandwidth. We propose RingFed to overcome these two problems.

\begin{algorithm}[t]
	\caption{Federated Averaging (FedAvg)}
	\label{alg1}
	\footnotesize
	\KwIn{Initial parameters $\textbf{w}_{init}$, probability $\mathscr{P}$, communication round $T$, local epoch $E$, learning rate $\eta$, partitioned non-IID data $\mathcal{D}.$} 
	
	\KwOut{Updated parameters $\textbf{w}$.} 

	\For{t=0:T-1}{
	
	\textbf{Clients}:
	
	\textbf{Each client $k$ in parallel}
	
	\Begin{ 
	
	    \If{$t=0$}{
            $\textbf{w}^t_k=\textbf{w}_{init}$ 
	    }
	
		\For{e=0:E-1}{
			a client updates parameters by the Stochastic Gradient Descent (SGD), and then $\mathbf{w}_k^{t,(e+1)}=\mathbf{w}_k^{t,e}-\eta \mathbf{g}(\mathbf{w}_k^{t,e};\zeta_k^e)$
		}
		Each client uploads $\mathbf{w}_k^{t,E}$ to the server
		}
	
		\textbf{Server}:
	    
	    \Begin{
	   
	    Aggregates the uploaded parameters $\mathbf{w}^{t+1}=\frac{1}{K}\sum_{m=0}^K\mathbf{w}_{m}^{t,E}$
	    
	    Downloads parameters $\mathbf{w}$ to all the $\mathcal{K}$ clients, $\mathbf{w}=\mathbf{w}^{t+1}$
	    
	    \If{$t \ne T-1$}{Selects $K$ clients from all the clients $\mathcal{K}$ with $\mathscr{P}$ probability in uniform distribution
        $K=\mathcal{K} \times \mathscr{P}$, $\mathscr{P} \in (0,1)$}
	   	}
    	}
\end{algorithm}

\begin{algorithm}[!t]
	\caption{RingFed}
	\label{alg2}
	\footnotesize
	\KwIn{Initial parameters $\textbf{w}_{init}$, probability $\mathscr{P}$, communication round $T$, local epoch $E$, learning rate $\eta$, and partitioned non-IID data $\mathcal{D}$, period $P$, and exchange factor $\gamma$.} 
	
	\KwOut{Updated parameters $\textbf{w}$.} 
	
	\For{t=0:T-1}{\tcp{A Communication Round Begins}
	
	\textbf{Clients}:
	
	\textbf{Each client $k$ in parallel}
	
	\Begin{ 
	    
	    \If{$t=0$}{
            $\textbf{w}^t_k=\textbf{w}_{init}$ 
	    }
	    \For{p=0:P-1}{
				\tcp{A Period Begins}
		\For{e=0:E-1}{
		    \tcp{An Epoch Begins}
			A client updates parameters by the Stochastic Gradient Descent (SGD), and then $\mathbf{w}_k^{t,p,e+1}=\mathbf{w}_k^{t,p,e}-\eta \mathbf{g}(\mathbf{w}_k^{t,p,e};\zeta_k^{t,p,e})$
		    }
		    \For{k=0:K-1}{
					clients exchange parameters with $\gamma \in [0,1]$ 
					$\mathbf{w}_{k+1}^{t,p,e+1}=\gamma\mathbf{w}_k^{t,p,e+1}+(1-\gamma)\mathbf{w}_{k+1}^{t,p,e+1}$
				}
				$\mathbf{w}_{0}^{t,p,e+1}=\gamma\mathbf{w}_{K}^{t,p,e+1}+(1-\gamma)\mathbf{w}_{0}^{t,p,e+1}$
		               
		}
		Each client uploads $\mathbf{w}_k^{t,P,E}$ to the server
		}
	
		\textbf{Server}:
	    
	    \Begin{
	    Aggregates the uploaded parameters $\mathbf{w}^{t+1,P,E}=\frac{1}{K}\sum_{k=0}^K\mathbf{w}_{k}^{t,P,E}$
	    
	    Downloads parameters $\mathbf{w}$ to all the $\mathcal{K}$ clients, $\mathbf{w}=\mathbf{w}^{t+1}$
	    
	    \If{$t \ne T-1$}{Selects $K$ clients from all the clients $\mathcal{K}$ with $\mathscr{P}$ probability in uniform distribution
        $K=\mathcal{K} \times \mathscr{P}$, $\mathscr{P} \in (0,1)$}
	   	}
    	}
\end{algorithm}
\subsection{Proposed Method: RingFed}

One factor of communication efficiency is the asymmetric property of internet connection speeds, i.e., the speed of uploading is lower than that of downloading \cite{Mcmahan2017Communication}. Moreover, some model compression algorithms are available to reduce downlink bandwidth usage \cite{Han2015Deep}, and the application of an encryption algorithm increases the number of parameters uploaded, aggravating the utilization efficiency of uplink bandwidth \cite{Bonawitz2017Practical}. Therefore, reducing uplink communication is feasible. Typically, FL uses a star topology, where each client communicates with only the central server, potentially increasing communication overhead when the global model is large. Inspired by the network topology, adjacent clients (e.g., those in the same institution) can be connected in a ring topology, facilitating the asymmetric property due to the proximity of clients, which increases the speed of uploading.

Note that after all the clients communicate with the central server, the process of aggregation begins, which causes communication congestion when uploading parameters within a short period of time. Moreover, a precipitate aggregation of averaging all participating clients worsens the global model, resulting in drastic oscillation, which we demonstrate in experiments. One question to consider is whether it is operative to “preaverage” by exchanging parameters between clients for times of training before uploading them to the central server. Based on this idea, we propose a novel optimization method that, in contrast to conventional methods, where clients communicate only with the server, directly changes the communication structure, making it possible for clients to preaggregate with each other to achieve better prediction accuracy and reduce the number of communication rounds from clients to the central server.

Clients are close physically in the local area network, which can reduce communication delays and accelerate upstream and downstream transmission rates in practice. A central server is also introduced to assist in the passing of parameters from training epochs, rather than being used as a training server for the model, as shown in Fig. \ref{fig:label_1}.
After $E$ epochs in local parameter updating by \eqref{SGD}, an additional step exchanging parameters between clients is performed  as follows 
\begin{equation}
	\left\{
	\begin{aligned}
		& \mathbf{w}_{k+1}=\gamma\mathbf{w}_k+(1-\gamma)\mathbf{w}_{k+1}, & k=0, 1, \cdots, K-1,\\
		& \mathbf{w}_{0}=\gamma\mathbf{w}_0+(1-\gamma)\mathbf{w}_{K},
	\end{aligned}
	\right.
\end{equation}
where $\gamma$ controls the degree of combination of parameters. Each client is indexed by the server to ensure that it does not exchange repeatedly. Then, another E epochs are conducted until the exchange period is completed. The total number of exchange parameters is termed period P. The concrete steps of RingFed are expressed in Algorithm \ref{alg2}, which further explains when a communication round, a period and an epoch begin.

The main difference between RingFed and other algorithms, as detailed in section \ref{section4} is that RingFed adds a step of recalculating the parameters for each client, which yields some excellent advantages. If $\gamma$ selects the number $\frac{1}{2}$ or $1$, each client absorbs parameters from another client to varying degrees as its initial network parameters for the next training period. In this way, after a certain number of periods, the model on each can learn features from different datasets without accessing the original data, mitigating the impact of the non-IID distribution to some extent. Moreover, when the set number of periods is reached, the client sends the parameters to the central server, greatly reducing the communication network consumption. Furthermore, RingFed can achieve fast convergence flatly and reach lower loss values, which indicates that the model predicts well on the test dataset. In addition, RingFed is equivalent to FedAvg if $\gamma=0$. 

\begin{table*}[tbp]
\centering
\caption{Datasets and Models Used in Experiments.}
\label{table_datasets}
\renewcommand\arraystretch{1.1}
\begin{tabular}{ccccccc}
\toprule[1pt]
\textbf{Dataset} & \textbf{No.Classes} & \textbf{Training Images} & \textbf{Testing Images} & \textbf{Total Images} & \textbf{Description}                                                                             & \textbf{Model}                                                              \\ \midrule
MNIST            & 10                  & 60,000                   & 10,000                  & 70,000                & Grayscale handwritten digits                                                                    & \multirow{2}{*}{LeNet}                                                      \\
FMNIST           & 10                  & 60,000                   & 10,000                  & 70,000                & Grayscale fashion products                                                                      &                                                                             \\ \midrule
EMNIST           & 47                  & 112,800                  & 18,800                  & 131,600               & \begin{tabular}[c]{@{}c@{}}Gray-scale handwritten digits,\\ uppercase and lowercase letters\end{tabular} & \multirow{2}{*}{\begin{tabular}[c]{@{}c@{}}Simplified\\ VGG\end{tabular}} \\
CIFAR-10         & 10                  & 50,000                   & 10,000                  & 60,000                & Color images of common animals and vehicles                                                               & 
\\ \bottomrule[1pt]
\end{tabular}
\end{table*}

\section{Experiments}\label{section4}
In this section, we present simulations to illustrate the advances of the proposed method compared to 3 other FL optimization methods \cite{Mcmahan2017Communication, Li2018Federated, Karimireddy2020Scaffold}. 

\subsection{Settings}

\subsubsection{Datasets and Neural Networks}or extensive validation, we compare our proposed method with three other methods on four different datasets and construct different neural networks for each dataset. The information of all datasets is summarized in Table \ref{table_datasets}.

\textbf{MNIST.} The MNIST dataset has 28×28 grayscale images of the numbers zero to nine, including 60,000 training images and 10,000 test images \cite{Lecun1998Gradient}.  

\textbf{Fashion-MNIST.} Fashion-MNIST comprises 28 × 28 grayscale images of 70,000 fashion products from 10 categories, including T-shirts, dresses, bags, etc. \cite{Xiao2017Fashion}.

\textbf{EMNIST.} The EMNIST dataset extends the MNIST with handwritten uppercase and lowercase letters and allows the selection of different models to classify the data. In our experiments, we choose the balanced mode, which contains 112,800 training images, 18,800 test images, and up to 47 classes \cite{Cohen2017Emnist}.

\begin{figure}[t]
	\centering
	\includegraphics[scale=0.27, trim=70 80 0 150]{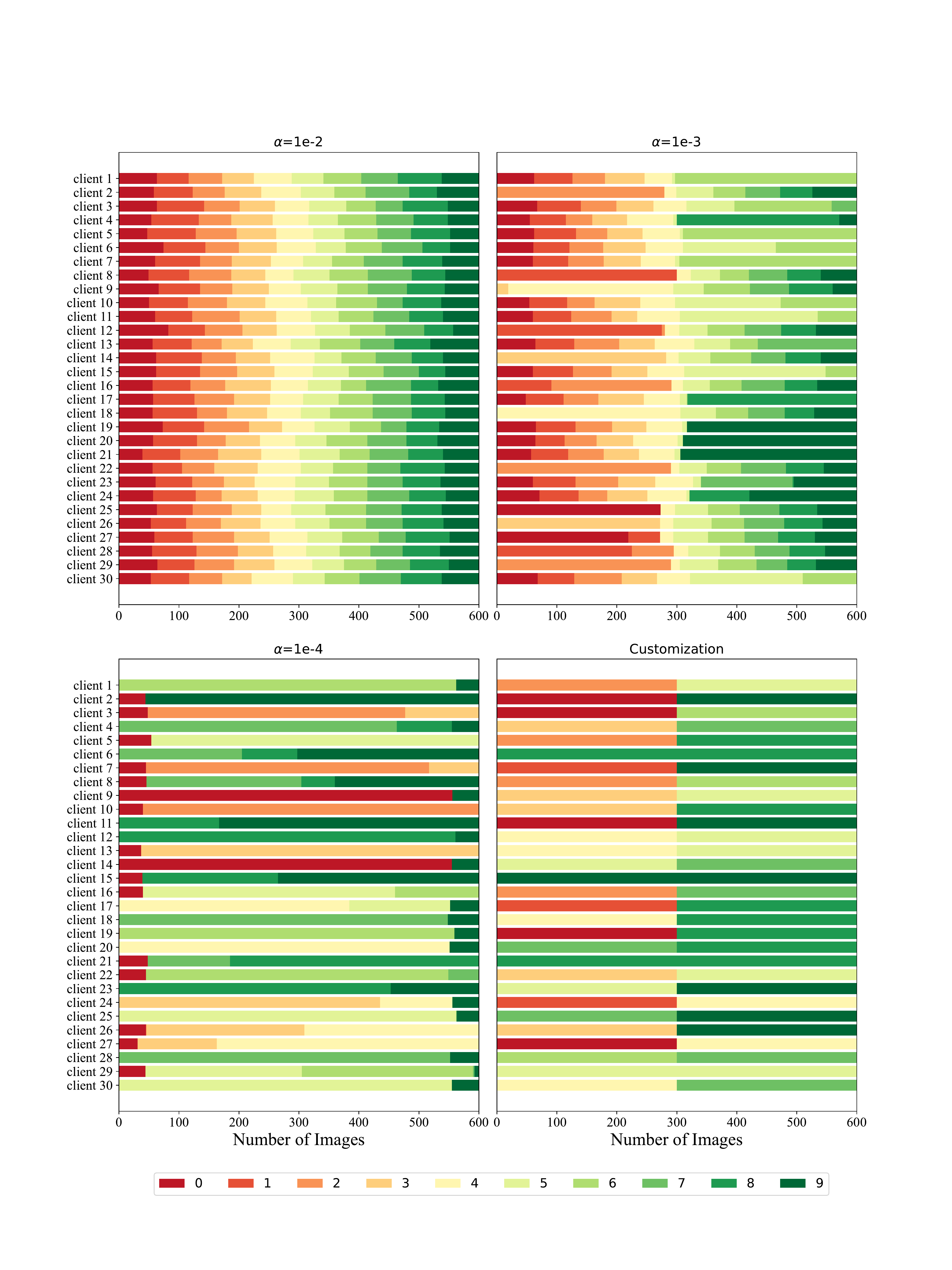}
	\caption{Visualization of synthetic data on MNIST. Thirty clients are randomly selected from all clients, and each row of a subgraph represents the data owned by one client, with different colors indicating different labels in the dataset.}
	\label{fig_dir}
\end{figure}

\textbf{CIFAR-10.} The CIFAR-10 dataset contains 10 common animal and vehicle species and consists of 50,000 training and 10,000 testing 32×32 color images \cite{Krizhevsky2009Learning}.

\textbf{Models.} In this experiment, we use two types of convolutional neural networks to classify images due to the varying complexity of the datasets. For the relatively simple MNIST and FMNIST, we train on LeNet with 2 convolutional layers and 3 fully connected layers. For the EMNIST and CIFAR-10 datasets, we simplify VGG11 by reducing the filters to [16, 32, 64, 64]. Moreover, all the batch normalization layers are removed due to poor performance with small batch size and non-IID data \cite{Sattler2019Robust}.

\subsubsection{Implementation} 
To more realistically simulate the FL environment and to ensure a fair comparison between the various methods, we make the following implementations.

\begin{figure*}[tp]
	\centering
	\includegraphics[scale=0.43, trim=70 0 0 200]{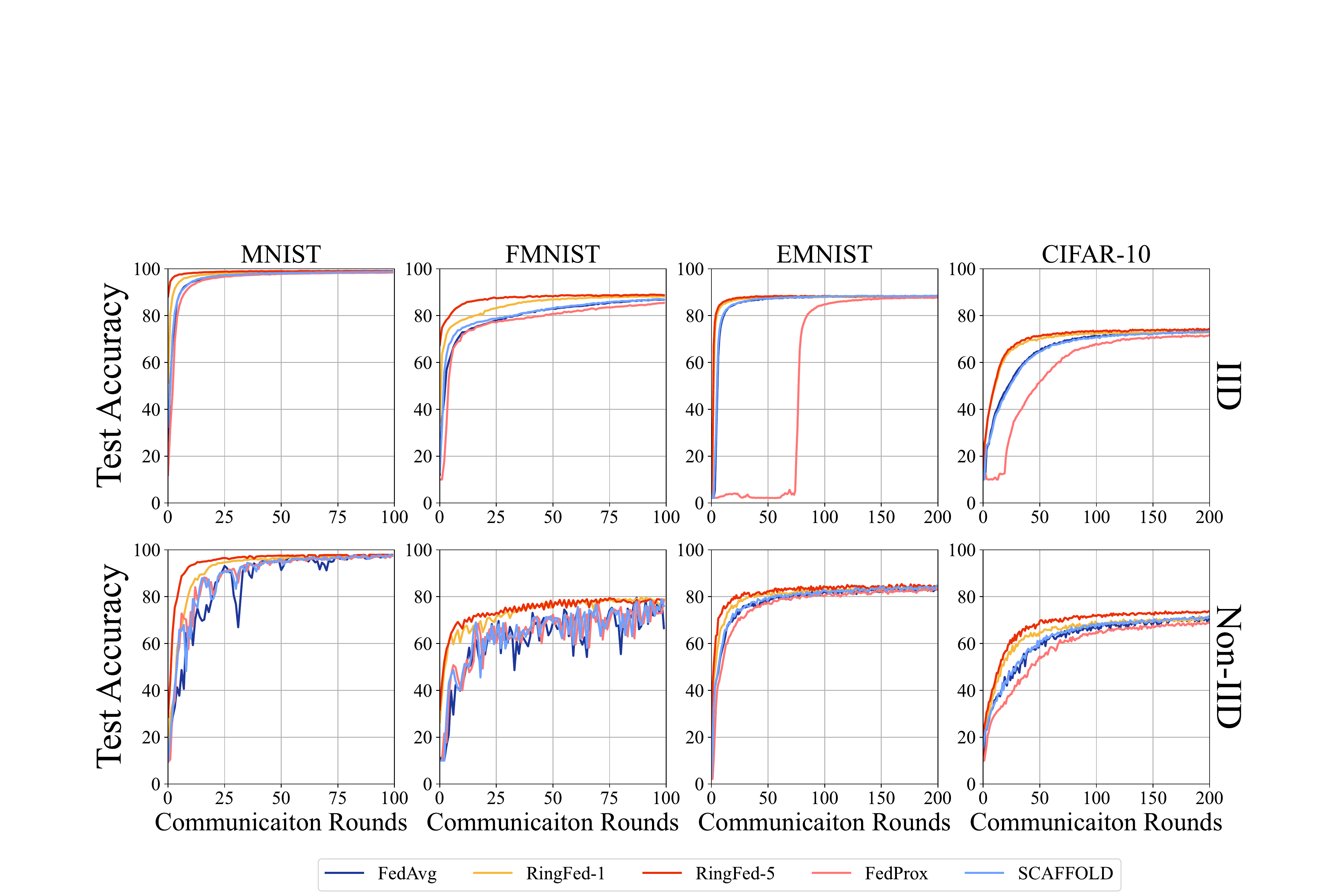}
	\caption{Test accuracy vs. communication rounds in different data distributions.}
	\label{fig_IID}
\end{figure*}

\begin{figure}[tbp]
	\centering
	\includegraphics[scale=0.32,trim=30 0 0 70]{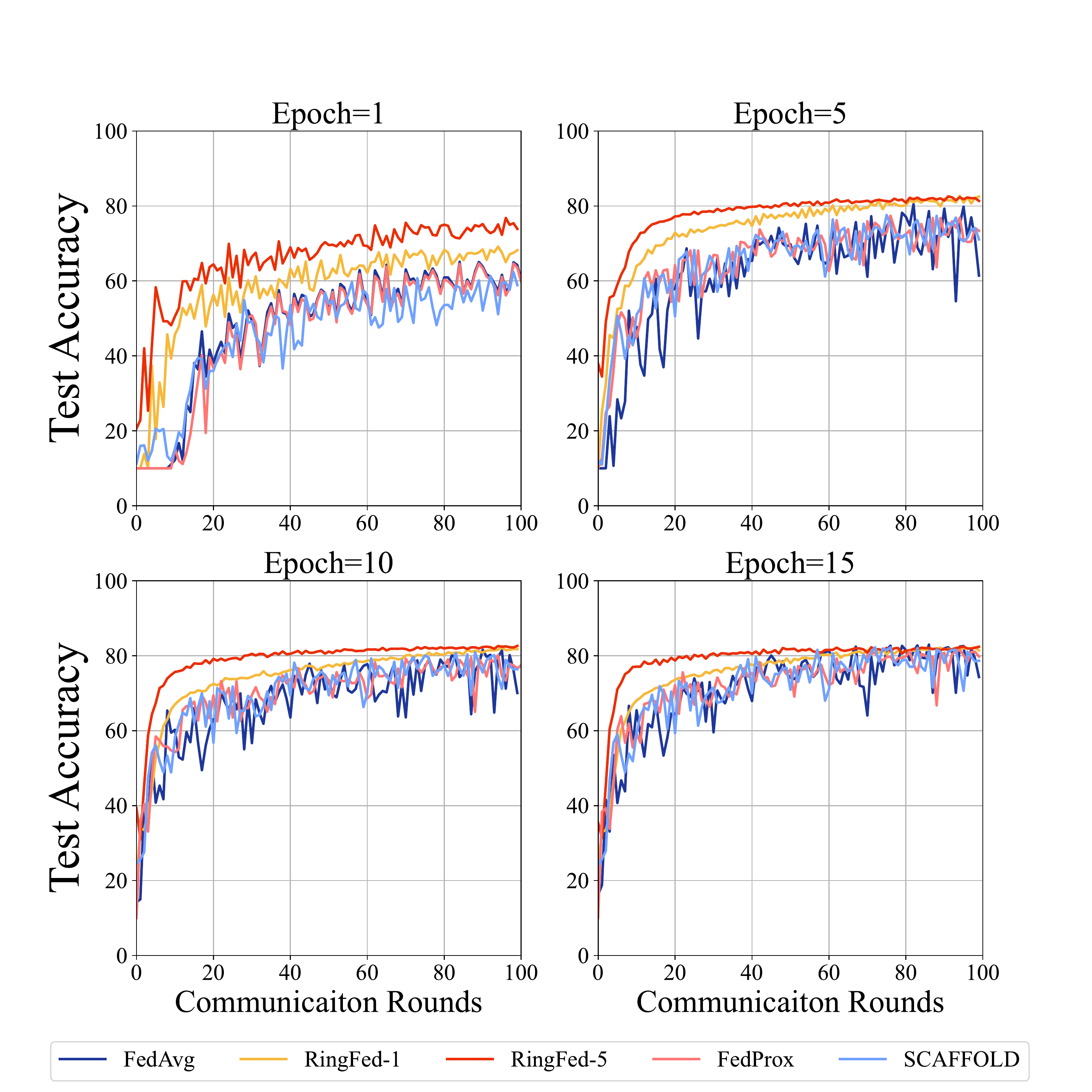}
	\caption{Test accuracy vs. communication rounds in different epochs.}
	\label{fig_epochs}
\end{figure}

\textbf{Non-IIDness.} 
One of the greatest challenges in FL is non-IID data, which makes it difficult for global models to converge and lowers the accuracy. We transform the distribution of the datasets in the following way.

Each client is allocated the same amount of data, i.e., each client has an amount of data equal to the number of training images for each dataset divided by the total number of clients $\mathcal{K}$ which is fixed to 100 in all the following experiments. Furthermore, for the MNIST and FMNIST datasets, we sort the data by digit and fashion product labels into 200 shards with 300 images and assign each of 100 clients 2 shards. This is a pathological non-IID partition of the data, as most clients will have examples of only two digits or products. Thus, this approach allows us to observe the degree to which algorithms will fail on highly non-IID data \cite{Mcmahan2017Communication}. However, for more complex datasets, pathological characterization is extremely disastrous for the performance of the algorithms. To synthesize data with fewer non-IID characteristics, we sample data for each client from the Dirichlet distribution $p\sim Dir(\alpha), \alpha>0$ \cite{Hsu2019Measuring}, where $\alpha$ is the concentration parameter controlling the Non-IID characteristic of the data assigned to clients, as shown in Fig \ref{fig_dir} taking the MNIST as an example. We finally choose the value of $\alpha$ as 0.001.

\textbf{Initial parameters and hyperparameters.} To ensure a fair comparison among algorithms, we set all conditions as equal as possible, such that only the algorithms differ. Without accounting for stragglers that cannot complete the training task within the specified time due to the low computing capability of the device in the actual scenario, all clients are simulated using uniform GPUs to guarantee consistent client performance. Additionally, we set random seeds to ensure that the following situations are also aligned among the 4 algorithms in terms of running code.

\begin{figure*}[tp]
	\centering
	\includegraphics[scale=0.45,trim=50 60 0 100]{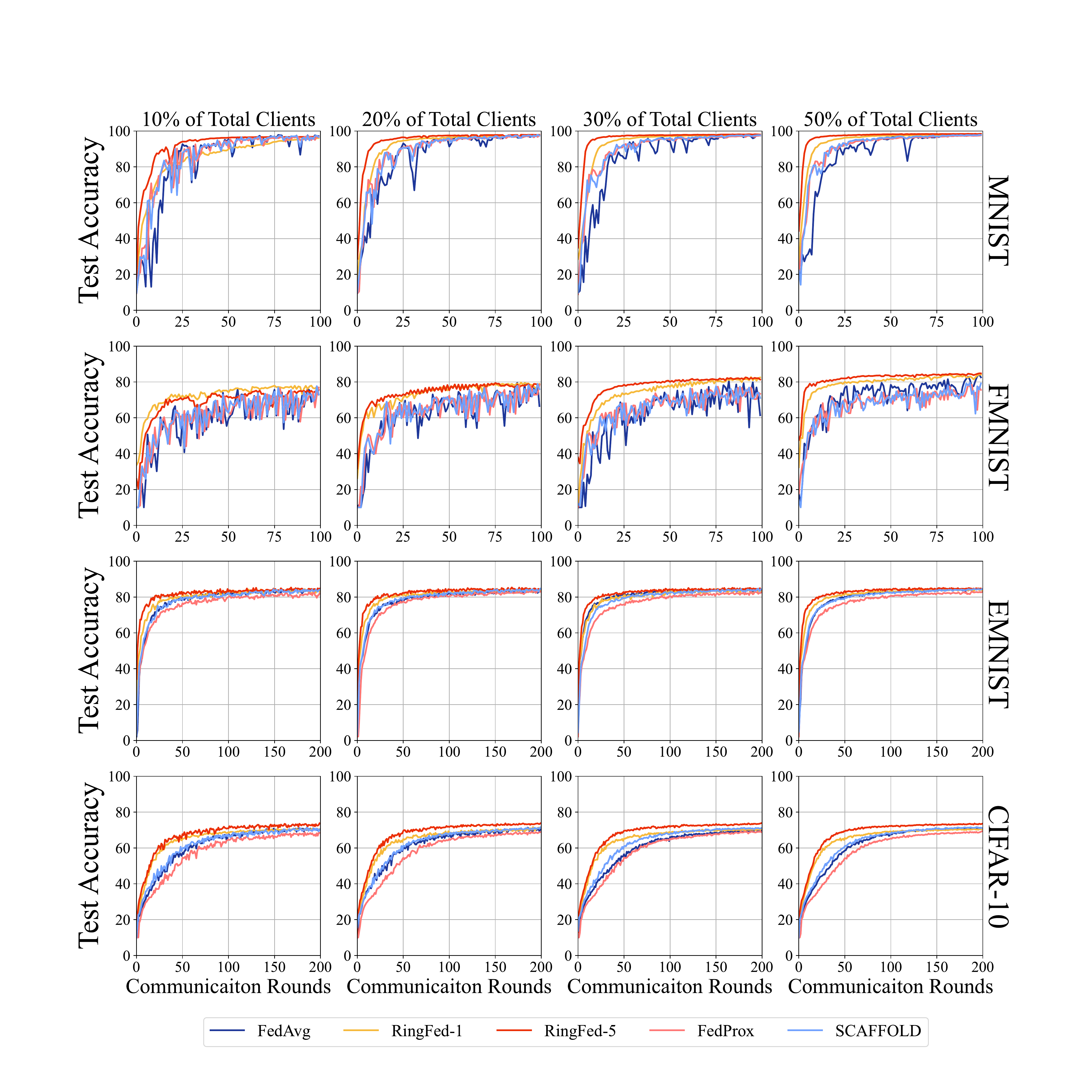}
	\caption{Test accuracy vs. diverse selected clients per round using for each datasets.}
	\label{fig_num_clients}
\end{figure*}

We set 100 clients in total and use the IID and transformed non-IID datasets to train the model. Furthermore, RingFed-1 means period $P=2$, while the RingFed-5 means the period $P=5$ where multiple exchanges are made between clients.

\subsection{Simulation Results}

\subsubsection{Preliminary experiments}We run the preliminary experiments with LeNet on the MNIST and FMNIST datasets and with a simplified VGG net on the EMNIST and CIFAR-10 datasets mentioned previously. For the IID setting, we spilt the original uniform training data randomly into equal shards. Each client is assigned a shard, and the number of shards depends on the size of the training set and the total number of clients. For the non-IID setting, the process is described in detail in Section \ref{section4}, Subsection A. Additionally, 30\% of the clients are selected for training in each round, and all experimental environments and parameters are set to be consistent, except for the distribution characteristics of the data.

Fig \ref{fig_IID} shows the impact of different data distributions on the global model performance. With respect to the IID data, due to the sophisticated convolutional neural network approach, all these algorithms are able to reach convergence quickly and smoothly despite the FL environment. However, for non-IID data, the convergence of the global model becomes more difficult, and large fluctuations in overall performance are observed, resulting in model instability. In addition, the final accuracy achieved by the model is reduced for the same number of communication rounds. Furthermore, the degree of model fluctuation and the reduction in accuracy are related to the strength of the non-IID characteristics of the data, as indicated by comparing $\alpha=0.001$ and the pathological non-IID partition with the IID data.
 
\begin{table*}[htbp]
\centering
\caption{Performance on Different Clients and Datasets}
\label{tab_num_clients}
\renewcommand\arraystretch{1.3}
\begin{threeparttable}
\begin{tabular}{ccccccccccccccccc}
\toprule[1pt]
\multirow{2}{*}{\textbf{Dataset}} & \multirow{2}{*}{\textbf{\%}} & \multicolumn{3}{c}{\textbf{FedAvg}} & \multicolumn{3}{c}{\textbf{RingFed-1}} & \multicolumn{3}{c}{\textbf{RingFed-5}} & \multicolumn{3}{c}{\textbf{FedProx}} & \multicolumn{3}{c}{\textbf{SCAFFOLD}} \\ \cmidrule(r){3-5} \cmidrule(r){6-8} \cmidrule(r){9-11} \cmidrule(r){12-14} \cmidrule(r){15-17} 
                                  &                              & RND\tnote{1}   & Acc.\tnote{2}            & C.C.\tnote{3}   & RND          & Acc.            & C.C.  & RND          & Acc.            & C.C.  & RND       & Acc.        & C.C.       & RND       & Acc.        & C.C.      \\ \midrule
\multirow{4}{*}{MNIST}            & 10                           & 48    & \textbf{97.82}   & 1        & 82           & 96.12           & 1.71  & \textbf{29}  & 96.77           & \textbf{0.6}   & 51        & 96.98       & 1.06       & 55          & 97.46       & 1.15      \\
                                  & 20                           & 42    & 97.88            & 1        & 28           & 97.5            & 0.67  & \textbf{16}  & \textbf{97.95}  & \textbf{0.38}  & 43        & 97.51       & 1.02       & 43          & 97.91       & 1.02      \\
                                  & 30                           & 38    & 97.65            & 1        & 22           & 97.81           & 0.58  & \textbf{8}   & \textbf{98.18}  & \textbf{0.21}  & 42        & 97.5        & 1.11       & 34          & 97.72       & 0.89      \\
                                  & 50                           & 43    & 97.8             & 1        & 19           & 98.05           & 0.44  & \textbf{7}   & \textbf{98.46}  & \textbf{0.16}  & 36        & 97.57       & 0.84       & 31          & 97.89       & 0.72      \\ \hline
\multirow{4}{*}{FMINST}           & 10                           & 55    & 76.33            & 1        & \textbf{45}  & \textbf{77.98}  & \textbf{0.82}  & 65           & 75.56           & 1.18  & 78        & 76.69       & 1.42       & 77          & 77.68       & 1.4       \\
                                  & 20                           & 69    & \textbf{79.57}   & \textbf{1}        & 39           & 79.73           & 0.52  & \textbf{29}  & 79.42           & \textbf{0.42}  & 70        & 78.48       & 1.01       & 61          & 78.92       & 0.88      \\
                                  & 30                           & 54    & 80.51            & 1        & 34           & \textbf{82.74}  & 0.63  & \textbf{14}  & 82.56           & \textbf{0.26}  & 62        & 77.42       & 1.15       & 61          & 77.6        & 1.13      \\
                                  & 50                           & 25    & 84.17            & 1        & 10           & 84.12           & 0.4   & \textbf{4}   & \textbf{84.79}  & \textbf{0.16}  & 54        & 78.57       & 2.16       & 56          & 79.64       & 2.52      \\ \hline
\multirow{4}{*}{EMNIST}           & 10                           & 61    & 84.56            & 1        & 41           & 84.47           & 0.67  & \textbf{20}  & \textbf{85.33}  & \textbf{0.33}  & 89        & 82.72       & 1.46       & 62          & 84.42       & 1.02      \\
                                  & 20                           & 58    & 84.32            & 1        & 33           & 84.41           & 0.57  & \textbf{22}  & \textbf{85.37}  & \textbf{0.38}  & 74        & 83.32       & 1.28       & 55          & 84.65       & 0.95      \\
                                  & 30                           & 37    & 84.52            & 1        & 34           & 84.47           & 0.92  & \textbf{22}  & \textbf{85.18}  & \textbf{0.59}  & 73        & 82.92       & 1.97       & 55          & 84.68       & 1.49      \\
                                  & 50                           & 50    & 84.46            & 1        & 36           & 84.47           & 0.72  & \textbf{18}  & \textbf{85.15}  & \textbf{0.36}  & 68        & 83.06       & 1.36       & 52          & 84.45       & 1.04      \\ \hline
\multirow{4}{*}{CIFAR-10}         & 10                           & 77    & 70.97            & 1        & 48           & 70.73           & 0.62  & \textbf{35}  & \textbf{73.76}  & \textbf{0.45}  & 103       & 68.64       & 1.34       & 72          & 71.36       & 0.94      \\
                                  & 20                           & 72    & 71.36            & 1        & 50           & 71.69           & 0.69  & \textbf{32}  & \textbf{73.73}  & \textbf{0.44}  & 100       & 68.85       & 1.39       & 69          & 71.5        & 0.96      \\
                                  & 30                           & 87    & 70.05            & 1        & 47           & 70.76           & 0.54  & \textbf{32}  & \textbf{73.8}   & \textbf{0.37}  & 96        & 69.32       & 1.1        & 66          & 71.2        & 0.76      \\
                                  & 50                           & 78    & 71.39            & 1        & 45           & 70.61           & 0.58  & \textbf{30}  & \textbf{73.66}  & \textbf{0.38}  & 94        & 69.25       & 1.21       & 63          & 71.62       & 0.81     
\\ \bottomrule[1pt]
\end{tabular}
\begin{tablenotes}    
        \footnotesize               
        \item[1] Number of communication rounds taken to reach a certain value of accuracy for the first time. 90\% for  MNIST, 75\% for FMNIST, 80\% for EMNIST, 65\% for CIFAR-10.      
        \item[2] The maximum accuracy that can be achieved in all communication rounds. 
        \item[3] Compared to FedAvg termed as 1, the amount of communication costs.
      \end{tablenotes}            
\end{threeparttable}
\end{table*}
 
\subsubsection{Effect on Non-IID Data with Different Parameters}
\textbf{Changing computation.} Changing computation. We first study the effects of the number of epochs on FMNIST. We fix the number of clients per round of sampling at 30 (100 in total) and gradually increase the number of epochs by five per trial from 5 to 15. Specifically, we run an experiment that sets 10 epochs for FedAvg, FedProx, and SCAFFOLD, 5 epochs for RingFed-1 and 2 epochs for RingFed-5, ensuring that the same amount of training is performed for all three methods. The results are shown in Fig. \ref{fig_epochs}.
We observe that (1) in the first three experiments, RingFed allows for faster and smoother convergence, which indicates that RingFed communicates with the central server less to achieve the same accuracy. (2) When the number of epochs is gradually increased, the accuracy of the global model increases, but the improvement is not significant. Moreover, increasing the number of epochs makes the training task for each client heavier and can even result in overfitting. Therefore, a larger number of epochs is not preferable. (3) Regardless of the training computation, RingFed converges smoothly, demonstrating robustness to the non-IID properties of the dataset.

\begin{table}[tbp]
\centering
\caption{Hyperparameters  Matrix}
\label{tab_hypers}
\begin{tabular}{cc}
\toprule[1pt]
\textbf{Hyperparameter} & \textbf{Values}        \\ \midrule
Learning rate           & 1e-4, 5e-4, 1e-3, 5e-3 \\
Momentum                & 0.9, 1.0               \\
Learning rate decay     & 0.98, 0.99, 1.0       \\ \bottomrule[1pt]
\end{tabular}
\end{table}

\textbf{Changing selected clients.} The hyperparameters of CNN can greatly influence the effectiveness of the model in practice, especially relative to SGD. To cope with this situation, a parameter matrix summarized in Table \ref{tab_hypers} is created. By letting different algorithms perform extensive experiments in combinations of the matrix parameters, we select the relatively best performance of each algorithm for comparison.

We fix the number of epochs $E=5$ and change the selected clients per round to study the influence of the two algorithms. We select 10\%, 20\%, 30\%, and 50\% of the clients, i.e., 10, 20, 30, and 50 random clients in each communication round, respectively, on 4 different datasets to provide parameters to form the global model. In addition, we calculate the test accuracy after each round via the global model and graph the results of the experiment, as shown in Fig. \ref{fig_num_clients}. We observe that (1) as the number of clients selected in each communication round increases, more local models are involved in the aggregation process. Therefore, the global model is able to achieve higher accuracy. (2) With the enhanced complexity of the four dataset image features and the deterioration of the degree of non-IID characteristics, the local models trained at each client have greater variation, resulting in drastic divergence of FedAvg, FedProx, and SCAFFOLD. In contrast, RingFed converges smoothly under all circumstances. (3) Additionally, RingFed achieves the same accuracy with fewer communication rounds, thereby reducing the communication costs. We summarize the comparative results in Table \ref{tab_num_clients}. 

\textbf{Changing $\gamma$.}
\begin{figure}[tbp]
	\centering
	\includegraphics[scale=0.33,trim=50 100 0 70]{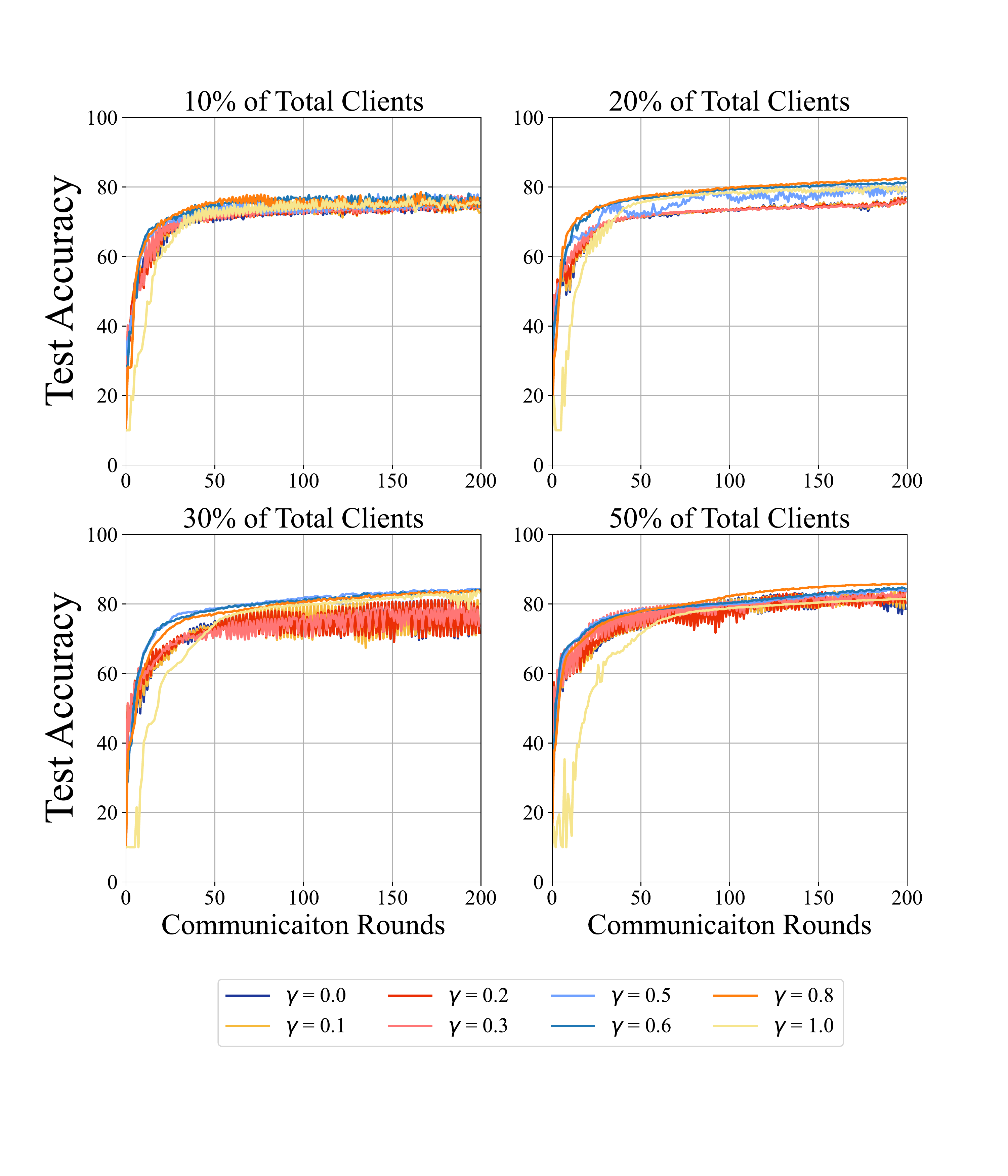}
	\caption{Test accuracy vs. communication rounds in different number of selected clients to compare the effects of various $\gamma$s.}
	\label{fig_gamma}
\end{figure}
The parameter $\gamma$ represents represents the extent to which a client’s parameters change in a period. If $\gamma$ is small, in each period, the parameters are exchanged, and they are less affected by the other parameters that are passed, while they are more affected when $\gamma$ is larger. Specifically, if $\gamma=0$, RingFed degenerate to the FedAvg; if $\gamma=1$, the parameters of a client are completely replaced by those of another client. To probe more deeply into the effect of $\gamma$ on RingFed, we fix epoch $E=5$, period $P=2$ and select $\gamma=0.0, 0.1, 0.2, 0.3, 0.5, 0.6, 0.8, 1.0$ respectively on FMNIST with 10\%, 20\%, 30\%, 50\% of total clients. The results are illustrated in Fig. \ref{fig_gamma}. In this experiment, interestingly, we find that if $0<\gamma<0.5$, the performance of RingFed deteriorates in terms of both convergence rate and test accuracy. If $0.5 \leq \gamma \leq 1.0$, the method converges rapidly and smoothly, with the best performs when $\gamma$ equals 0.8 instead of 1.0. 
 we summarize the average accuracy and variance for different values when the global model tends to be stable in Table \ref{table_2}.

\begin{table}[ht]
	\centering
	\caption{Statistics of different $\gamma$ in the last 50 communication rounds.}
	\label{table_2}
	\begin{tabular}{ccccc}
		\toprule[1pt]
		\textbf{\% of clients}           & \textbf{$\gamma$} & \textbf{Mean}             & \textbf{Stdev}              \\
		\midrule
		\multirow{8}{*}{10} & 0(FedAvg)                 & 74.46                     & 1.13                     \\
		& 0.1               & 74.54                     & 1.22                     \\
		& 0.2               & 74.71                     & 1.26                     \\
		& 0.3               & 75.21                     & 1.53                     \\
		& 0.5               & 75.64                     & 1.61                     \\
		& 0.6               & \textbf{75.83}                     & 1.51                     \\
		& 0.8               & 75.49                     & 1.16                     \\
		& 1.0               & 75.22                     & 0.93                     \\
		\midrule
		\multirow{8}{*}{20} & 0(FedAvg)                 & 74.98                     & 0.87                     \\
		& 0.1               & 75.01                     & 0.78                     \\
		& 0.2               & 74.99                     & 0.61                     \\
		& 0.3               & 74.75                     & 0.52                     \\
		& 0.5               & 78.82                     & 1.10                     \\
		& 0.6               & 80.86                     & \textbf{0.24}                     \\
		& 0.8               & \textbf{81.83}                     & 0.41                     \\
		& 1.0               & 79.37                     & 0.62                     \\
		\midrule 
		\multirow{8}{*}{30} & 0(FedAvg)                 & 75.94                     & 4.99                     \\
		& 0.1               & 75.43                     & 4.29                     \\
		& 0.2               & 76.07                     & 4.77                     \\
		& 0.3               & 76.62                     & 2.34                     \\
		& 0.5               & \textbf{83.73}                     & 0.40                     \\
		& 0.6               & 83.27                     & \textbf{0.37}                     \\
		& 0.8               & 83.23                     & \textbf{0.37}                     \\
		& 1.0               & 82.21                     & 1.01                     \\
		\midrule
		\multirow{8}{*}{50} & 0(FedAvg)                 & 80.92                     & 1.63                     \\
		& 0.1               & 81.42                     & 1.75                     \\
		& 0.2               & 82.17                     & 1.50                     \\
		& 0.3               & 81.79                     & 1.09                     \\
		& 0.5               & 83.32                     & 0.57                     \\
		& 0.6               & 83.78                     & 0.57                     \\
		& 0.8               & \textbf{85.44}                     & \textbf{0.27}                     \\
		& 1.0               & 81.00                     & 0.30 \\
		\bottomrule[1pt]
	\end{tabular}
\end{table}

\section{Conclusion}\label{section5}
In this paper, we propose a federated learning optimization algorithm called RingFed, which transforms the topology of communication network so that clients can communicate with each other. Our experimental results show that RingFed outperforms FedAvg in most cases, achieving higher test accuracy in fewer communication rounds and faster convergence. Additionally, we optimize the results of training on non-IID data and further reduce the communication cost. Future work will focus on extensions to broad datasets and deeper neural networks.

\bibliographystyle{IEEEtran}

\bibliography{IEEEabrv,IEEEexample}

\end{document}